\PassOptionsToPackage{prologue,dvipsnames}{xcolor}
\documentclass[sigconf,natbib=true,anonymous=false]{acmart}

\usepackage{threeparttable}
\usepackage{multirow}
\usepackage{makecell}
\usepackage{color, soul}
\usepackage{tcolorbox}
\usepackage{colortbl}
\usepackage[utf8]{inputenc}
\usepackage{enumitem}
\usepackage{amsthm}
\usepackage{amsmath}
\usepackage{bm}

\newtheorem{definition}{Definition} 
\AtBeginDocument{%
  }

\setcopyright{acmlicensed}
\copyrightyear{2018}
\acmYear{2018}
\acmDOI{XXXXXXX.XXXXXXX}

\acmConference[Conference acronym 'XX]{Make sure to enter the correct
  conference title from your rights confirmation emai}{June 03--05,
  2018}{Woodstock, NY}
\acmISBN{978-1-4503-XXXX-X/18/06}

\begin{document}


\title[Disentangling Instructive Information from Ranked Multiple Candidates]{Disentangling Instructive Information from Ranked Multiple Candidates for Multi-Document Scientific Summarization
}

\author{Pancheng Wang}
\orcid{0000-0002-4450-8848}
\affiliation{%
  \institution{National University of Defense Technology}
  \city{Changsha}
  \country{China}
}
\email{wangpancheng13@nudt.edu.cn}

\author{Shasha Li}
\authornote{Corresponding authors.}
\orcid{0000-0002-6508-5119}
\affiliation{%
  \institution{National University of Defense Technology}
  \city{Changsha}
  \country{China}
}
\email{shashali@nudt.edu.cn}

\author{Dong Li}
\affiliation{%
  \institution{National University of Defense Technology}
  \city{Changsha}
  \country{China}
}
\email{lidong1@nudt.edu.cn}

\author{Kehan Long}
\affiliation{%
  \institution{National University of Defense Technology}
  \city{Changsha}
  \country{China}
}
\email{longkehan15@nudt.edu.cn}

\author{Jintao Tang}
\authornotemark[1]
\orcid{0000−0002−8802−3906}
\affiliation{%
  \institution{National University of Defense Technology}
  \city{Changsha}
  \country{China}
}
\email{tangjintao@nudt.edu.cn}

\author{Ting Wang}
\authornotemark[1]
\orcid{0000-0002-7780-2330}
\affiliation{%
  \institution{National University of Defense Technology}
  \city{Changsha}
  \country{China}
}
\email{tingwang@nudt.edu.cn}








\renewcommand{\shortauthors}{Trovato et al.}

\begin{abstract}
Automatically condensing multiple topic-related scientific papers into a succinct and concise summary is referred to as Multi-Document Scientific Summarization (MDSS). Currently, while commonly used abstractive MDSS methods can generate flexible and coherent summaries, the difficulty in handling global information and the lack of guidance during decoding still make it challenging to generate better summaries. To alleviate these two shortcomings, this paper introduces summary candidates into MDSS, utilizing the global information of the document set and additional guidance from the summary candidates to guide the decoding process. Our insights are twofold: Firstly, summary candidates can provide instructive information from both positive and negative perspectives, and secondly, selecting higher-quality candidates from multiple options contributes to producing better summaries. Drawing on the insights, we propose a summary candidates fusion framework --- \textbf{D}isentangling \textbf{I}nstructive information from \textbf{R}anked candidates (\textbf{DIR}) for MDSS. Specifically, DIR first uses a specialized pairwise comparison method towards multiple candidates to pick out those of higher quality. Then DIR disentangles the instructive information of summary candidates into positive and negative latent variables with Conditional Variational Autoencoder. These variables are further incorporated into the decoder to guide generation. We evaluate our approach with three different types of Transformer-based models and three different types of candidates, and consistently observe noticeable performance improvements according to automatic and human evaluation. More analyses further demonstrate the effectiveness of our model in handling global information and enhancing decoding controllability.
  
\end{abstract}




\maketitle
\section{INTRODUCTION}
\label{sec1}

\textbf{M}ulti-\textbf{D}ocument \textbf{S}cientific \textbf{S}ummarization (\textbf{MDSS}) involves integrating dispersed information across multiple topic-related scientific articles and generating a succinct and concise summary. To address this task, current abstractive methods \cite{jin2020multi, chen2022target, wang2022multi} utilize the input scientific articles as context information and generate the summary with a Transformer-based decoder. Such abstractive methods can generate flexible and coherent output, but they have two drawbacks: Firstly, MDSS takes multiple scientific articles with diverse and abundant information as input, which makes it difficult for the decoder to accurately grasp the global context of the input and focus on precise input context at each decoding step. Secondly, although MDSS belongs to the task of conditional text generation, the output summaries still have various discourse structures and content selections without violating the input documents. However, the decoder only relies on the input documents as conditional guidance for decoding, making it challenging to pre-select and plan the summary content, resulting in difficulties in controlling the content of the summary.

Previous research has attempted to address the shortcomings in the decoder and enhance decoding performance. For instance, Chen et al. \cite{chen2021capturing} model the cross-document relationships in scientific articles and use relationships to guide content selection in the decoder. Wang et al. \cite{wang2022multi} model entities and relationships in the input documents and focus on these entities and relationship information during decoding. The above works facilitate the decoding process by mining features within input documents, which can provide further control and guidance for summary generation, compared with relying merely on the input context. However, strategies designed solely based on the input documents themselves are still suboptimal choices, because when the decoder is powerful enough, it may implicitly exploit potential features among multiple documents to achieve better decoding. Therefore, the motivation of this paper is to introduce external guidance information to facilitate content selection and summary decoding process, which aims to achieve superior performance than the summarization guided by mining internal information.

A high-quality external guidance information that can achieve the above objectives is \textbf{summary candidates}. Here, summary candidates specifically refer to the summaries generated by abstractive methods for the current input. The advantages of using summary candidates for the second-stage summarization include: (i) Summary candidates integrate salient information and global semantic information from multiple documents in advance, which can help the decoder achieve global control over the document set and converge the attention during decoding; (ii) Summary candidates can provide additional guidance information for summary generation, helping the decoder decide which content should be emphasized and which content should be avoided during decoding.
 
Previous studies using summary candidates for summarization fall into two categories: (i) \textbf{Ranking-based methods} involve re-ranking several candidates to pick the one with the highest quality \citep{liu2021simcls,ravaut2022summareranker}. However, these methods lack the ability to integrate information from multiple candidates to generate a new summary, thereby limiting their potential to enhance decoder performance. (ii) \textbf{Fusion-based methods} concatenate multiple summary candidates directly after the input documents to serve as context information for decoding \citep{ravaut2022towards,jiang2023llm}. While these methods show some improvement, we contend that mere concatenation cannot fully exploit the advantages of summary candidates and fails to address the issue of controlling summary content during decoding.

Therefore, in this paper, we explore a more elaborate way to
incorporate summary candidates into MDSS based on the following two motivations: (i)\textbf{ Motivation 1: Summary candidates can provide instructive information from both positive and negative perspectives} (\S{\ref{sec3.1}}). The positive instructive information should be enhanced, while the negative instructive information should be suppressed when generating the second-stage summaries. (ii) \textbf{Motivation 2: Ranking and selecting higher-quality candidates from a pool of multiple candidates helps to produce better second-stage summaries} (\S{\ref{sec3.2}). Higher quality indicates the contents of the candidates are more informative and instructive. Selecting multiple summary candidates allows for the comprehensive integration of guidance information from different candidates, maximizing the guidance effect of summary candidates.

Based on the motivations, in this paper, we devise a candidates fusion framework --- \textbf{D}isentangling \textbf{I}nstructive information from \textbf{R}anked candidates (\textbf{DIR}), which is compatible with various Transformer-based MDSS models and includes three modules:
\vspace{-0.3em}
\begin{itemize}[leftmargin=*]
\item {\textit{Candidates Ranking Module}}: We utilize a candidates ranking module to pick out higher quality candidates for subsequent candidates fusion. Specifically, the ranking module is a specialized pairwise comparison method, designed to discern subtle differences between candidates and enhance ranking performance.
\item{\textit{Instructive Information Modeling and Disentangling Module}}: We model positive and negative information of summary candidates with separate latent variables. Each latent variable is trained by a Conditional Variational Autoencoder (CVAE) \citep{kingma2013auto,sohn2015learning}. To make sure positive and negative aspects are disentangled from each other, we employ two auxiliary training objectives: an informativeness objective and an adversarial objective to enhance the disentanglement.

\item{\textit{Information Augmented Decoding Module}}: We design the information augmented decoding module under the Transformer-based encoder-decoder architecture in two ways: (i) Extending the source context by concatenating the representations of candidates with those of source papers; (ii) Modifying the decoder layer by injecting positive and negative latent variables.
\end{itemize}

In our work, we consider three types of first-stage models to produce the summary candidates: Large Language Model\footnote{\url{https://platform.openai.com/docs/models/gpt-3-5}}, KGSum \citep{wang2022multi}, and BART \citep{lewis2020bart}. These models exhibit diverse capabilities and generate summaries in different ways. We integrate the candidates fusion framework DIR into three Transformer-based MDSS models: TransS2S \citep{wang2022multi}, KGSum \citep{wang2022multi}, and BART \citep{lewis2020bart}, and we conduct experiments across three large-scale MDSS datasets: Multi-Xscience \citep{lu2020multi}, TAD \citep{chen2022target} and TAS2 \citep{chen2022target}. The experimental results show that our DIR framework brings substantial improvements on multiple base models across the three datasets. The results highlight that selecting higher-quality candidates and disentangling candidate information into positive and negative aspects are two effective strategies to improve the performance of fusion-based summarization. Our code is available at: \url{https://github.com/muguruzawang/DIR}.

The main contributions of this work are summarized below:
\begin{itemize}[leftmargin=*]
\item We introduce summary candidates to enhance the decoder's handling of global information and provide decoding guidance information for MDSS.
\item We propose a candidates fusion framework DIR, which is compatible with various Transformer-based model. DIR includes three modules: \textit{Candidates Ranking}, \textit{Instructive Information Modeling and Disentangling}, and \textit{Information Augmented Decoding}. 
\item We incorporate the proposed DIR framework into three models and experiment on three MDSS datasets, and the results indicate the effectiveness of the DIR framework.
\end{itemize}


\vspace{-1em}
\section{Related Work}
\subsection{Multi-Document Scientific Summarization}
Multi-Document Scientific Summarization (MDSS) involves consolidating scattered information from multiple papers. According to the paradigm adopted, previous studies can be categorized into graph-based \citep{chen2022target,wang2022multi}, flat-based \citep{moro2022discriminative,shi2023towards}, and hierarchical-based methods \citep{chen2021capturing,shen2023hierarchical}. Concretely, graph-based methods construct external graphs (e.g., entity-relation graph, AMR, TF-IDF) to assist document representation and cross-document relation modeling. In this regard, \citet{chen2022target} leverage graphs to encode source articles, constructing keyphrase graphs and using a multi-level contrastive learning strategy to improve the accuracy of generated summaries. \citet{wang2022multi} incorporate knowledge graphs into document encoding and decoding, generating the summary from a knowledge graph template. Flat-based methods simply concatenate multiple documents, thus they can leverage state-of-the-art pre-trained summarization models. \citet{moro2022discriminative} select top $K$ documents with a dense retriever and use a BART model \citep{lewis2020bart} to produce the summary. \citet{shi2023towards} leverage Large Language Model (LLM) to expand the query for reference retrieval and adopt an instruction-tuning method to guide LLM to generate summary.  Compared with flat-based methods, hierarchical-based methods take the hierarchical structure of document clusters into account, thus preserving cross-document relations and obtaining semantic-rich representations. \citet{chen2021capturing} apply word representations and cross-document relationships to model document-level representations, and design a hierarchical decoding strategy. \citet{shen2023hierarchical} modify the BART model by incorporating document-level representations in the encoder and decoder to facilitate multi-document interactions. 

However, we argue that all the above works fail to explore external guidance information for better global semantic control and summary decoding.

\vspace{-1em}
\subsection{Summarization with Summary Candidates}
Previous studies \citep{liu2021simcls,ravaut2022summareranker,liu2022brio} introduce candidates into summarization to address the exposure bias problem \citep{ranzato2016sequence}, which refers to the discrepancy between training time and inference time for generation tasks with maximum likelihood estimation. The usual paradigm is to train a re-ranker to select the higher-quality candidates. In this respect, \citet{liu2021refsum} use meta-learning to re-rank summary candidates produced by several base systems. \citet{liu2021simcls} train a RoBERTa to re-rank candidates with contrastive learning and a ranking loss, while \citet{ravaut2022summareranker} train a binary classifier as re-ranker with binary cross-entropy loss. Further, \citet{liu2022brio} re-use the base model for a second round of fine-tuning with both the cross-entropy loss and a candidate-level ranking loss. Although effective, re-ranking based methods are bounded by the quality of the first-stage candidates.

In view of the above drawback, \citet{ravaut2022towards} propose a fusion-based method to combine multiple candidates. They concatenate the representations of multiple candidates with those of source papers. Similar fusion strategy also exists in \citep{jiang2023llm}, where they concatenate the text together. We contend that mere concatenation fails to fully harness the information within the candidates, which motivates us to consider more effective approaches to fuse candidates.

\vspace{-0.5em}
\section{Motivation}
In this section, we elaborate on our motivations for how to utilize summary candidates.
\vspace{-0.5em}
\subsection{Candidates: Positive vs Negative?}
\label{sec3.1}
Summary candidates are generated by the first-stage summarization models. These candidates, as the predicted condensed version of the document set, may align with certain aspects of the gold summary while also exhibiting mismatches in some aspects, as shown in Figure \ref{pos_neg}.
\begin{figure}[th]
  \centering
  \vspace*{-0.3\baselineskip}
   \includegraphics[width=0.95\columnwidth]{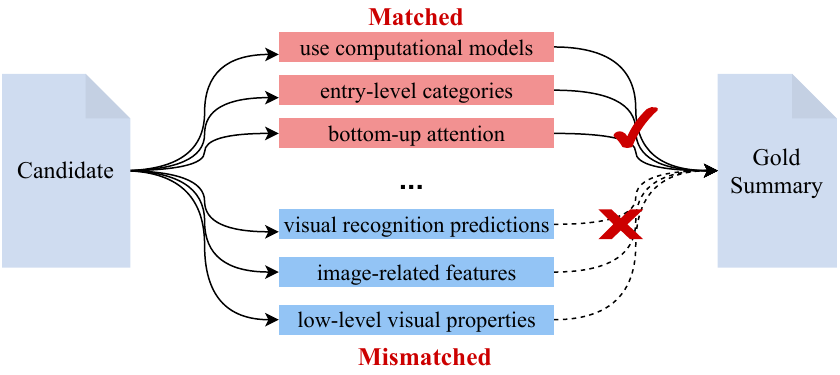}
   \vspace*{-0.5\baselineskip}
     \caption{The summary candidate may match the gold summary in some aspects and not in others.}
     \vspace*{-0.5\baselineskip}
    \label{pos_neg}
\end{figure}
We argue that matched contents can serve as positive signals to encourage the model to generate such contents, whereas the mismatched contents as negative signals to discourage the model from generating those contents. 


To effectively represent the positive and negative information of candidates, we consider an intuitive approach --- using \textbf{unigram set} based concept to  define them.
\vspace*{-0.4\baselineskip}
\begin{definition}
We define the \textbf{Positive Information} $I_p$ to be the \textbf{intersection} between the  unigram set of the candidate ($cand$) and that of the gold summary ($gold$): ${I_p} = {C^{uni}}(cand) \cap {C^{uni}}(gold)$, where ${C^{uni}}(x)$ means the unigram set of $x$.
\vspace*{-0.4\baselineskip}
\end{definition}

\vspace*{-0.4\baselineskip}
\begin{definition}
We define the \textbf{Negative Information} $I_n$ to be the \textbf{difference set} between the unigram set of the candidate  ($cand$) and that of the gold summary ($gold$): ${I_n} = {C^{uni}}(cand) - {C^{uni}}(gold)$, where ${C^{uni}}(x)$ means the unigram set of $x$.
\vspace*{-0.4\baselineskip}
\end{definition}

The rationality of the above definitions is twofold: Firstly, unigram-based metric is the widely recognized measure of machine translation \citep{papineni2002bleu} and summarization \citep{lin2004rouge}, indicating that unigram-based approach is effective in capturing the similarity between the candidate and the gold summary. Hence, it is reasonable and intuitive to model $I_p$ and $I_n$ using the intersection and difference of unigram sets; Secondly, based on the above definitions, it is feasible to model $I_p$ and $I_n$ using latent variables and enhance the informativeness of $I_p$ and $I_n$ with Bag-of-Words prediction.

\vspace{-0.5em}
\subsection{Is Selecting Higher-quality Candidates Beneficial?}
\label{sec3.2}

While both positive and negative information of candidates could be instructional for second-stage summarization, our preference leans towards incorporating more positive information, i.e., more overlapped unigram. This is because, in the most extreme scenario, the gold summary as the candidate consistently outperforms random text as the candidate. Based on this premise, we aim to figure out whether re-ranking multiple candidates and selecting higher-quality candidates would be beneficial for this purpose. 

\begin{figure}[h]
  \centering
  \vspace*{-0.3\baselineskip}
   \includegraphics[width=0.99\columnwidth]{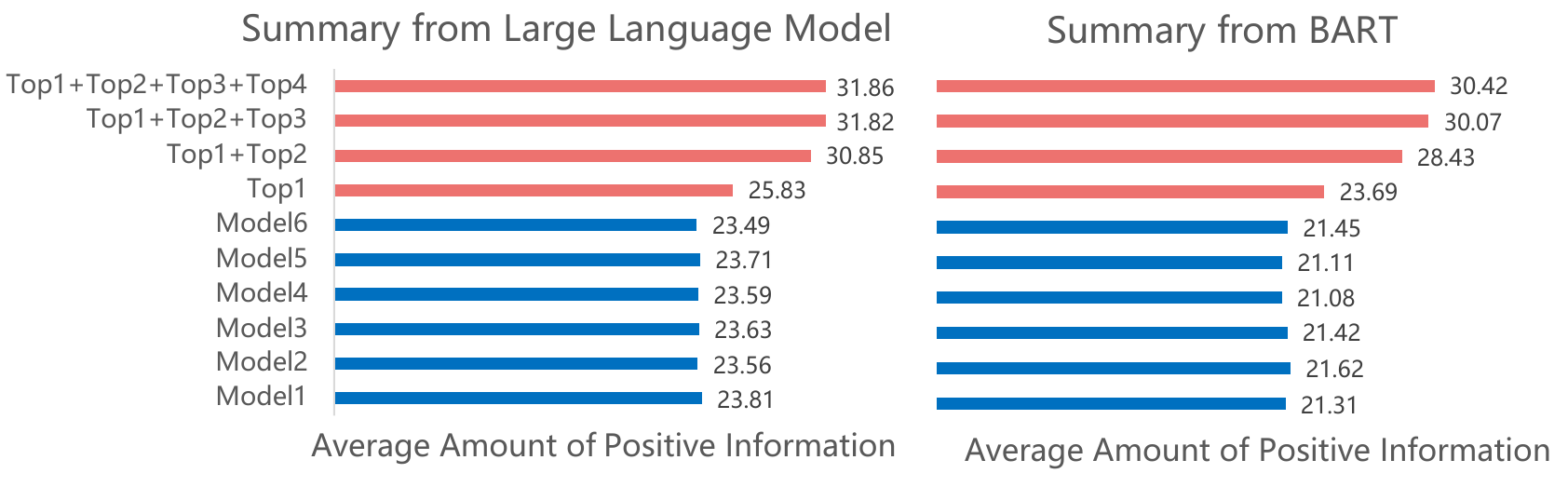}
   \vspace*{-0.3\baselineskip}
     \caption{The average amount of positive information of outputs from different first-stage models and selected top-$\{1,2,3,4\}$ candidates.}
    \label{pi}
    \vspace*{-0.7\baselineskip}
\end{figure}

To this end, we conduct an investigation on one MDSS dataset --- Multi-Xscience \citep{lu2020multi}. We construct the pool of multiple candidates with two types of first-stage models: Large Language Model (LLM)\footnote{\url{https://platform.openai.com/docs/models/gpt-3-5}} and BART \citep{lewis2020bart}. We interact with the LLM with six different prompts, thus obtaining six summary candidates for each test instance. Furthermore, we fine-tune BART and use six different training checkpoints to obtain another six summary candidates for each test instance. We re-rank the six candidates for each instance according to ROUGE-1 metric \citep{lin2004rouge}, which is a standard measure to assess the summary informativeness, in line with our purpose. \textbf{We aim to investigate the extent to which re-ranking multiple candidates can enhance the positive information within candidates.} The result is shown in Figure \ref{pi}.

 \begin{figure*}[th]
  \centering
   \includegraphics[width=0.9\textwidth]{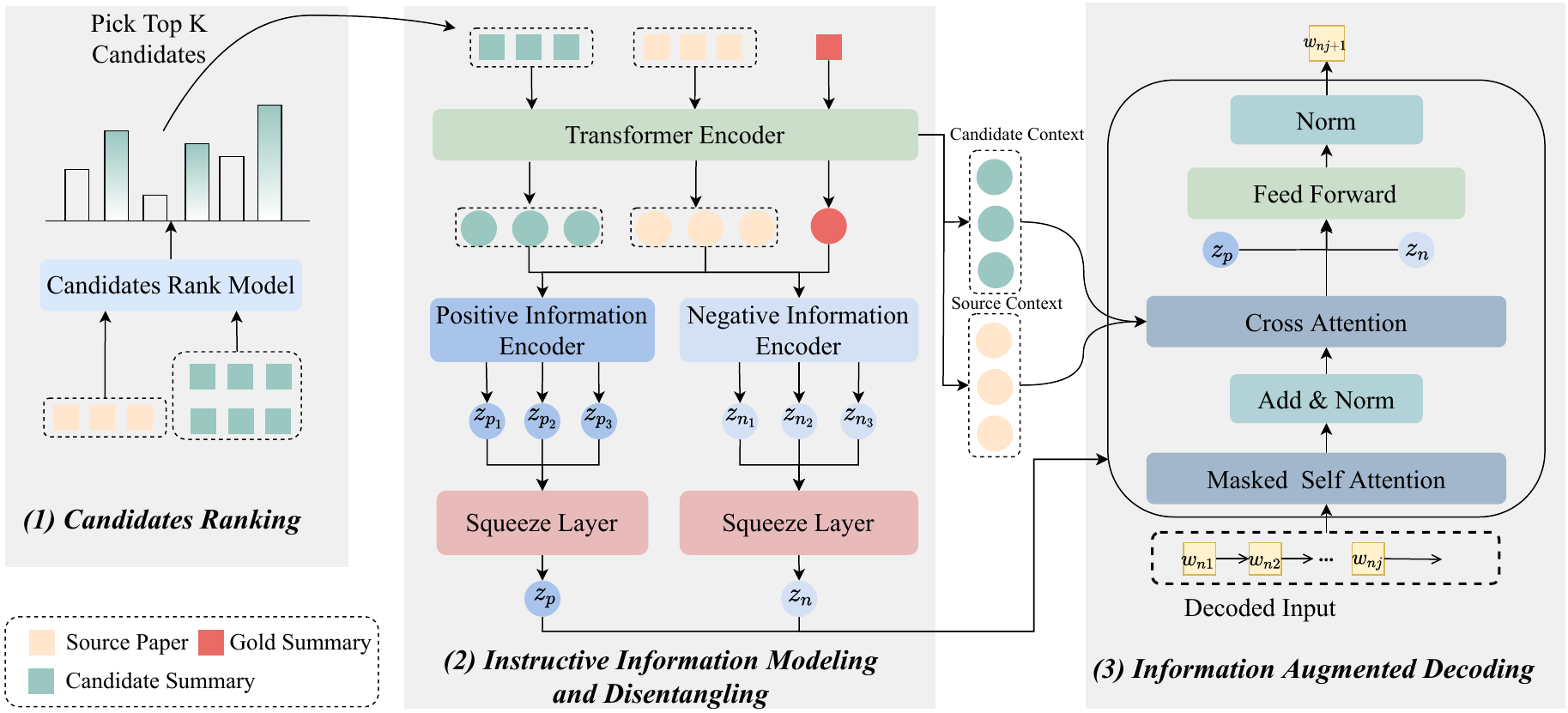}
   \vspace*{-0.3\baselineskip}
     \caption{The overall framework of our proposed Disentangling Instructive information from Ranked candidates (DIR).}
    \label{model}
    \vspace*{-0.5\baselineskip}
\end{figure*}

From Figure \ref{pi}, we find that the top-1 candidate has more positive information than any of the six models, which suggests that re-ranking multiple candidates and selecting the highest-quality candidate for each instance will be beneficial to improve the informativeness of positive information. What's more, it brings further improvements when combining top-K candidates together. Therefore, we decide to re-rank multiple candidates and select top-K candidates to take full advantage of the positive information.

\vspace*{-0.5\baselineskip}
\section{METHODOLOGY}

The framework of our proposed DIR is shown in Figure \ref{model}. DIR is a multiple candidates fusion framework that is compatible with any Transformer-based MDSS model. For convenience, we illustrate the modules of DIR based on the basic Transformer encoder-decoder architecture in Figure \ref{model}. DIR consists of three modules: \textit{Candidates Ranking} (\S{\ref{sec_cr}}), \textit{Instructive Information Modeling and Disentangling} (\S{\ref{sec_iird}}), and \textit{Information Augmented Decoding} (\S{\ref{sec_ird}}).

\vspace*{-0.5\baselineskip}
\subsection{Candidates Ranking}
\label{sec_cr}
The candidates ranking module (the left part of Figure \ref{model}) re-ranks multiple summary candidates according to ROUGE-1 metric. We frame the re-ranking problem as a binary classification, which has been successfully employed for re-ranking in previous work \citep{ravaut2022summareranker,jiang2023llm}. 

We need first to obtain a good representation of the summary candidate for classification. To enrich contextual information, we combine the representation of the source papers with that of the candidate to obtain the final representation. In view that source papers tend to have more than 1,000 tokens, we employ Longformer \citep{beltagy2020longformer} as the encoder. Concretely, we concatenate the source papers with special tokens as separators, resulting in the form of  \verb|<s><source paper 1><\s><source paper 2>...<source paper| \verb|n><\s>|. Then we send the source papers into Longformer, and the embedding of special token \verb|<s>| is used as the representation $h_{d}$ of source papers. Besides, each candidate is sent to Longformer individually to obtain the representation $h_c$ of the candidate. 

Once we have the representations of source papers and candidate $c_i$, we concatenate them together and use a two-layer feed-forward network to obtain the final prediction probability:
\vspace{-0.3em}
\begin{equation}
{p_\psi (c_i) } = \textrm{sigmoid}({w_2}{\rm{ReLU}}({w_1}[{h_{d}};{h_{c_i}}] + {b_1}) + {b_2})
\vspace{-0.3em}
\end{equation}
We optimize the candidates ranking module with a binary cross-entropy loss as follow:
\vspace{-0.3em}
\begin{equation}
{\mathcal{L}_{rank}} =  - {y_{{c_i}}}\log {p_\psi }({c_i}) - (1 - {y_{{c_i}}})\log (1 - {p_\psi }({c_i}))
\vspace{-0.3em}
\end{equation}
where $y_{c_i}=1$ if $c_i$ is the positive instance, otherwise $y_{c_i}=0$.

During training, to determine the positive and negative instances, we randomly select one of the top two ranked candidates as the positive instance, and randomly select one of the bottom two candidates as the negative instance.

It is worth noting that \textbf{the training of candidates ranking module is independent of the training of the other modules.} This is because, the candidates ranking module is designed to tackle the gap between the training and inference stages of the summarization model (\S{\ref{sec_iird}} and \S{\ref{sec_ird}}). To be specific, during the training stage of the summarization model, we have access to the oracle ranking of multiple candidates, which allows us to train the summarization model using the ground truth top-K candidates as input. However, during inference, the ranking of multiple candidates is not available because there is no gold summary. To remedy this gap, we use the trained candidates ranking module to predict the ranking result of candidates and select top-K candidates as input for summarization.

\vspace*{-0.5\baselineskip}
\subsection{Instructive Information Modeling and Disentangling}
\label{sec_iird}
In this subsection, we describe how to model the instructive information of summary candidates and disentangle it into positive and negative information.

\vspace{-0.5em}
\subsubsection{Text Encoding}
We first use Transformer Encoder to obtain text representations. As shown in the middle part of Figure \ref{model}, the encoder takes three types of inputs, $\{\mathcal{D}, \mathcal{C}, S\}$, representing the source papers, the candidates and the gold summary, respectively. Take the source papers $\mathcal{D} = \{ {d_1},{d_2},...,{d_{N_d}}\}$ for example, we add each source paper ${d_i}$ with special tokens, resulting in the form of  \verb|<s><source paper i><\s>|. We take the  \verb|<s>| representation from the last layer of the encoder as the representation of ${d_i}$, denoted as $h_{d_i}^E$. Then we apply \textit{mean-pooling} over $\{h_{d_i}^E\}$ to obtain the representation of the document set:
\vspace*{-0.4\baselineskip}
\begin{equation}
h_\mathcal{D}^E = \frac{1}{{{N_d}}}\sum\limits_{i = 1}^{{N_d}} {h_{{d_i}}^E}
\vspace*{-0.4\baselineskip}
\end{equation}
Similarly, we use the \verb|<s>| representation to obtain the representations of each summary candidate $h_{c_i}^E$, and the gold summary $h_g^E$. 

\vspace{-0.5em}
\subsubsection{Instructive Information Modeling}
We design two latent variational modules to model positive and negative information of candidates as independent latent variables. 

We use the Conditional Variational Autoencoder (CVAE) \citep{kingma2013auto,sohn2015learning} to achieve the goal. The CVAE introduces a prior distribution on the latent representation space $z$, typically represented as an isotropic Gaussian distribution. It also replaces the deterministic encoder with a learned approximation of the posterior distribution parameterized by a neural network. The training objective of CVAE includes two terms: a reconstruction loss that encourages the network to encode the information that is necessary to generate the output, and a KL divergence loss to push the approximate posterior distribution close to the prior distribution.

We map each selected summary candidate $c_i$ into two latent variables representing positive and negative information:  $z_{p_i}$ and $z_{n_i}$. They adopt the same neural network architecture. Take $z_{p_i}$ for example. We use isotropic Gaussian distribution as the prior distribution of $z_{p_i}$: 
${p_{\theta} }({z_{{p_i}}}|\mathcal{D},{c_i}) = {\mathcal{N}}(\mu_{p_i} ,{\sigma_{p_i} ^2}{\rm{I}})$, which means the prior distribution is conditioned on the source papers $\mathcal{D}$ and the candidate $c_i$, and $\rm{I}$ denotes the identity matrix. We apply a two-layer feed-forward network ($FFN$) to compute $\mu_{p_i}$ and $\sigma_{p_i} ^2$ as follows:
\vspace*{-0.1\baselineskip}
\begin{equation}
{\mu _{{p_i}}} = FFN([h_{\mathcal{D}}^E;h_{{c_i}}^E]), \log {\sigma _{{p_i}}^2} = FFN([h_{\mathcal{D}}^E;h_{{c_i}}^E])
\vspace*{-0.1\baselineskip}
\end{equation}
where $[;]$ means the concatenation operation.

At the training stage, the posterior distribution is conditioned not only on source papers $\mathcal{D}$ and candidate $c_i$, but also on gold summary $S$, which provides the necessary guidance for positive information modeling. The posterior distribution is defined as: ${q_\phi }(z_{{p_i}}^\prime|{\mathcal{D}},{c_i},S) = {\mathcal{N}}({\mu _{{p_i}}^\prime},\sigma _{{p_i}}^{\prime2}{\rm{I}})$, where $\mu_{p_i}^\prime$ and $\sigma_{p_i} ^{\prime2}$ are computed as:
\vspace*{-0.2\baselineskip}
\begin{equation}
{\mu _{{p_i}}^\prime} = FFN([h_{\mathcal{D}}^E;h_{{c_i}}^E;h_g^E]), \log {\sigma _{{p_i}}^{^\prime2}} = FFN([h_{\mathcal{D}}^E;h_{{c_i}}^E;h_g^E])
\vspace*{-0.2\baselineskip}
\end{equation}

After modeling the prior distribution and the posterior distribution, we can sample $z_{p_i}$ from the posterior and prior distribution during training and inference, respectively. The latent variable of negative information $z_{n_i}$ is modeled similarly to $z_{p_i}$.

\vspace{-0.5em}
\subsubsection{Instructive Information Disentangling}
Since we use the same input to generate $z_{p_i}$ and $z_{n_i}$, they are not distinguishable according to the training objective of CVAE described above. Therefore, we adopt two commonly used objectives for learning disentangled representations of positive and negative information.

\noindent\textbf{Informativeness Objective.} 
The latent variable should be informative enough to predict the corresponding generative factor. To achieve this objective, we use each latent variable to predict the Bag-of-Words (BOW) representation of the corresponding information. Concretely, for latent variable $z_{p_i}$ and its oracle BOW representation $y_{bow}^{p_i}$, the predicted BOW representation is calculated as:
\begin{equation}
\label{eq1}
{p_{inf }}({z_{{p_i}}}) = {\rm{softmax(}}{w_{inf }}{z_{{p_i}}} + {b_{inf }}{\rm{)}}
\end{equation}
Then the informativeness loss is computed with the cross-entropy loss as:
\vspace*{-0.4\baselineskip}
\begin{equation}
\label{eq7}
{\mathcal{L}_{inf}}(z_{{p_i}}) =  -\sum\limits_{j = 1}^{|bow|} {y_{bow}^{{p_i},j}\log ({p_{inf}}(z_{{p_i}}^j))}
\vspace*{-0.3\baselineskip}
\end{equation}
The informativeness loss of $z_{{n_i}}$ is calculated similarly to $z_{{p_i}}$.

\noindent\textbf{Adversarial Objective.}
Considering that one latent variable should be predictive of its corresponding information only, we train adversarial classifiers for each latent variable, following \citep{john2019disentangled}, which try to predict the information of the non-target generative factor. In this way, the model attempts to construct the latent spaces such that the non-target distributions predicted by these classifiers are as non-predictive as possible. 

The adversarial objective is composed of two parts. The first part is the adversarial classifiers on each latent variable for each type of non-target information. The second part is the loss aiming to maximize the entropy of the predicted distribution of the adversarial classifiers. Take the adversarial objective of $z_{{p_i}}$ for example, we train a linear classifier to predict the BOW representation of negative information. The loss of the adversarial classifier is computed as:
\vspace*{-0.3\baselineskip}
\begin{equation}
\label{eq8}
{{\mathcal{L}}_{adv}}({z_{{p_i}}}) =  - \sum\limits_{j = 1}^{|bow|} {y_{bow}^{{n_i},j}\log ({p_{adv}}(z_{{p_i}}^j))} 
\vspace*{-0.3\baselineskip}
\end{equation}
where ${{p_{adv}}(z_{{p_i}})}$ is the predicted output of the adversarial classifier, which is defined similarly to Equation \ref{eq1}. It is worth noting that the gradients of the adversarial classifiers are not back-propagated to the rest part of CVAE.

Additionally, we introduce another objective for each adversarial classifier, which aims to make its predicted distribution indistinguishable. We achieve this by maximizing the entropy of the predicted distribution ${{p_{adv}}(z_{{p_i}})}$:
\vspace*{-0.2\baselineskip}
\begin{equation}
\label{eq9}
{{\mathcal{H}}}({z_{{p_i}}}) =  - \sum\limits_{j = 1}^{|bow|} {{p_{adv}}(z_{{p_i}}^j)\log ({p_{adv}}(z_{{p_i}}^j))}
\vspace*{-0.2\baselineskip}
\end{equation}

The above informativeness objective and adversarial objective ensure that the latent variables are informative and disentangled from each other.

\vspace{-0.5em}
\subsubsection{Multiple Latent Variables Squeezing}
After obtaining $z_{{p_i}}$ and $z_{{n_i}}$ for each of the $k$ selected candidates, we introduce a squeeze layer, which obtains a unified representation that gathers the entire positive information or negative information across multiple candidates. Take $\{ {z_{{p_i}}}\} _{i = 1}^k$ for example, the squeezing operation is achieved by another $FFN$ network, taking the concatenation of $\{ {z_{{p_i}}}\} _{i = 1}^k$ as input:
\vspace*{-0.3\baselineskip}
\begin{equation}
{{\hat z}_p} = FFN([{z_{{p_1}}};{z_{{p_2}}};...;{z_{{p_k}}}])
\end{equation}
${\hat z}_n$ is obtained similarly to ${\hat z}_p$.

\vspace{-0.7em}
\subsection{Information Augmented Decoding}
\label{sec_ird}
We incorporate the information of summary candidates into decoding process in two ways (shown in the right part of Figure \ref{model}): (i) Extending the source context by concatenating the representations of candidates with those of source papers; (ii) Modifying the decoder layer by injecting positive and negative information.

\vspace*{-0.3\baselineskip}
\subsubsection{Source Context Concatenation}
To allow the decoder to focus on the candidates directly, we concatenate the encoding representations of candidate context with those of source context, which are obtained separately by the Transformer Encoder. Then the decoder performs cross-attention on the concatenated context:
\vspace*{-0.3\baselineskip}
\begin{equation}
{{h}}_{cross}^D{\rm{ = CrossAtt}}([{h_{source}};{h_{{c_1}}};...;{h_{{c_k}}}])
\vspace*{-0.3\baselineskip}
\end{equation}
where ${h_{source}}$ and ${h_{{c_i}}}$ denote the token-level representations of source papers and each candidate.
\vspace{-0.5em}
\subsubsection{Injecting positive and negative information}
To take full advantage of the disentangled positive and negative information, we incorporate the squeezed latent variables ${\hat z}_p$ and ${\hat z}_n$ into each decoder layer by modifying the cross-attention operation:
\begin{equation}
{{\hat h}}_{cross}^D = {{h}}_{cross}^D + {{\hat z}_p} - {{\hat z}_n}
\end{equation}

In this way, the decoder is taught to generate better output by enhancing or suppressing the influence of the instructive information of candidates. 

\vspace{-0.8em}
\subsection{Loss Function}
Our DIR framework can be incorporated into any Transformer-based summarization model. The loss function includes three terms: the negative evidence lower bound (ELBO) of CVAE ($\mathcal{L}_{ELBO}$), the loss of informativeness objective ($\mathcal{L}_{INF}$), and the loss of adversarial objective ($\mathcal{L}_{ADV}$).

Assuming that the number of selected summary candidates is $k$, $\mathcal{L}_{ELBO}$ is derived as:
\begin{equation}
    \begin{split}
{{\mathcal{L}}_{ELBO}} &=  - {\mathbb{E}_{{q_\phi }({z_{{p_{i:k}}}},{z_{{n_{i:k}}}}|\mathcal{D},\mathcal{C},S)}}[\log p(S|\mathcal{D},\mathcal{C},{z_{{p_{i:k}}}},{z_{{n_{i:k}}}})\\
 &+ \sum\nolimits_{i \in [1,k]} {{\rm{KL}}({q_\phi }({z_{{p_i}}}|\mathcal{D},{c_i},S)||{p_\theta }({z_{{p_i}}}|\mathcal{D},{c_i}))} \\
 &+ \sum\nolimits_{i \in [1,k]} {{\rm{KL}}({q_\phi }({z_{{n_i}}}|\mathcal{D},{c_i},S)||{p_\theta }({z_{{n_i}}}|\mathcal{D},{c_i}))} 
    \end{split}
\end{equation}

The second informativeness loss term $\mathcal{L}_{INF}$ is the summation of Equation \ref{eq7} across all latent variables:
\vspace*{-0.3\baselineskip}
\begin{equation}
{{\mathcal{L}}_{INF }} = \sum\nolimits_{i \in [1,k]} {{{\mathcal{L}}_{inf}}({z_{{p_i}}}) + {{\mathcal{L}}_{inf}}({z_{{n_i}}})}
\end{equation}

The third adversarial loss term $\mathcal{L}_{ADV}$ is the combination of Equation \ref{eq8} and \ref{eq9} across all latent variables:
\vspace*{-0.3\baselineskip}
\begin{equation}
{{\mathcal L}_{ADV}} = \sum\nolimits_{i \in [1,k]} {{{\mathcal L}_{adv}}({z_{{p_i}}}) + {{\mathcal L}_{adv}}({z_{{n_i}}})}  - {\mathcal H}({z_{{p_i}}}) - {\mathcal H}({z_{{n_i}}})
\vspace*{-0.5\baselineskip}
\end{equation}

Then we get the final loss function as:
\begin{equation}
{\mathcal L} = {{\mathcal L}_{ELBO}} + \lambda  \cdot {{\mathcal L}_{INF}} + \beta  \cdot {{\mathcal L}_{ADV}}
\end{equation}
where $\lambda$ and $\beta$ are hyper-parameter weights to be tuned on the validation set.

\section{EXPERIMENTS}
\vspace{-0.3em}
\subsection{Experimental Setup}
\subsubsection{Dataset}
We evaluate our method on three public large-scale MDSS datasets: Multi-Xscience \citep{lu2020multi}, TAD \citep{chen2022target} and TAS2 \citep{chen2022target}. Specifically, \textbf{Multi-Xscience} is collected from two sources: arXiv and the Microsoft Academic Graph (MAG). It contains 30,369/5,066/5,093 instances for training/validation/testing. Each instance contains the abstract of a query paper and the abstracts of reference papers it cites as input, and a paragraph in the related work section of the query paper as the gold summary. \textbf{TAD} and \textbf{TAS2} are collected from the public scholar corpora S2ORC \citep{lo2020s2orc} and Delve \citep{akujuobi2017delve}, respectively. TAD contains papers from multiple fields, while TAS2 focuses on computer science field. TAD and TAS2 contain 208,255/5,000/5,000 and 107,700/5,000/5,000 instances for training/validation/testing, respectively. The input and output formats of TAD and TAS2 are consistent with Multi-Xscience.

\vspace*{-0.5\baselineskip}
\subsubsection{Evaluation Metrics}
For the evaluation of MDSS, we consider both automatic and human evaluation, both of which are widely adopted in summarization. For automatic evaluation, we use ROUGE-based metric  \citep{lin2004rouge} to calculate the N-grams overlap between the output and gold summary.
We use ROUGE-1 (\textbf{R-1}), ROUGE-2 (\textbf{R-2}), and ROUGE-L (\textbf{R-L}) to take unigram, bigrams, and the longest common subsequence into consideration. Additionally, we use a semantic-based metric,  BERTScore (\textbf{BS}) \citep{zhang2019bertscore}, to compute the similarity score between the summaries based on their contextual embeddings.

We also conduct a human evaluation to provide a more comprehensive evaluation of the summary. We invite three well-educated Master student to do this job. The detailed settings of human evaluation is described in \S{\ref{sec5.4.2}}.

\vspace*{-0.5\baselineskip}
\subsubsection{Baselines}
To validate the effectiveness of our proposed DIR framework, we choose the following three types of baselines for comparison.

\textit{1) Extractive methods}:
\underline{TextRank} \citep{mihalcea2004textrank} and \underline{LexRank} \citep{erkan2004lexrank}: are two graph-based unspervised extractive models. \underline{HeterSumGraph}~\citep{wang2020heterogeneous}: a graph-based model based on heterogeneous graph, which incorporates semantic nodes of various levels of granularity. 

\textit{2) Abstractive methods}: \underline{Pointer-Generator} \citep{see2017get}: a seq2seq abstractive model based on Recurrent Neural Networks (RNN). \underline{MGSum} \citep{jin2020multi}: a multi-document summarization model that utilizes a multi-granularity interaction network. \underline{GraphSum} \citep{li2020leveraging}: 
a graph-based model that utilizes graphs to enhance both the encoding and decoding processes.
\underline{BertABS} \citep{liu2019text}: an abstractive model that utilizes BERT \citep{devlin2019bert} as the encoder and uses a randomly-initialized decoder. \underline{BART} \citep{lewis2020bart}: a transformer-based pretrained text generation model. \underline{PRIMERA} \citep{xiao2022primera}: a pretrained encoder-decoder model designed for multi-document summarization. \underline{TAG} \citep{chen2022target}: an abstractive model for MDSS, which leverages keyphrases graph and contrastive learning. \underline{$\mathrm{UR^3WG}$} \citep{shi2023towards}: an MDSS model that uses LLM to retrieve query and generate summary.
\underline{KGSum} \citep{wang2022multi}: a Transformer-based MDSS model that utilizes knowledge graph for encoding and decoding. \underline{EDITSum} \citep{WANG2023103409}: the current State-of-the-art model that leverages sentence-level planning for MDSS. \underline{TransS2S \citep{wang2022multi}}: a Transformer-based model with randomly-initialized encoder and decoder. 

\begin{table*}[t]
\caption{The performances of different models on MDSS across three datasets. The number of selected summary candidates $k=3$. ``\textit{Oracle}'' and ``\textit{Pred}'' mean we take the oracle ranking and the predicted ranking (obtained by \S{\ref{sec_cr}}) of candidates as input when testing, respectively. ``-'' means the result is inaccessible. ``$\dagger$'' and ``$\dagger\dagger$'' indicate statistically significantly better than the corresponding base model with Paired t-test $p<0.01$ and $p<0.05$. The underlined data refers to the previous SOTA.}
\vspace*{-0.5\baselineskip}
    \label{autoresult}
    \centering
    \renewcommand\arraystretch{0.7}
    \small
    \begin{threeparttable}
    \begin{tabular}{m{0.23\textwidth}<{\raggedright}|p{0.71cm}<{\centering}p{0.71cm}<{\centering}p{0.71cm}<{\centering}p{0.71cm}<{\centering}|p{0.71cm}<{\centering}p{0.71cm}<{\centering}p{0.71cm}<{\centering}p{0.71cm}<{\centering}|p{0.71cm}<{\centering}p{0.71cm}<{\centering}p{0.71cm}<{\centering}p{0.71cm}<{\centering}}
    \toprule
    \multirow{2}{*}{\textbf{Model}} & \multicolumn{4}{c|}{\textbf{Multi-Xscience (\%)}} & \multicolumn{4}{c|}{\textbf{TAD (\%)}} & \multicolumn{4}{c}{\textbf{TAS2 (\%)}} \\
    \cmidrule{2-13}
    & \textbf{R-1} & \textbf{R-2} & \textbf{R-L} & \textbf{BS} & \textbf{R-1} & \textbf{R-2} & \textbf{R-L}  & \textbf{BS} & \textbf{R-1} & \textbf{R-2} & \textbf{R-L}  & \textbf{BS} \\ \midrule
   \multicolumn{13}{l}{\emph{Extractive}}\\ \cmidrule{1-13}
    TextRank (\citeyear{mihalcea2004textrank})& 30.38& 5.30& 25.20 & 84.06 &  26.80 & 3.61& 22.39 & 83.11 & 26.19 & 3.14 & 21.39 & 83.45\\
    LexRank (\citeyear{erkan2004lexrank})& 29.52& 4.91& 25.32 & 84.04 & 28.57 & 4.52 & 24.80 & 83.71 &27.51& 3.77& 23.51 & 84.00\\
    HeterSumGraph (\citeyear{wang2020heterogeneous}) & 31.36& 5.82& 27.41 & - & 27.85& 3.88& 24.82 & -& 27.56 & 3.62& 24.25 & -\\\midrule
  \multicolumn{13}{l}{\emph{Abstractive}}\\\midrule
    MGSum (\citeyear{jin2020multi})& 33.11& 6.75& 29.43 & 85.76 & 27.49 & 4.79&  25.21 & -& 25.54 & 3.75 & 23.16& -\\
    GraphSum (\citeyear{li2020leveraging}) & 29.58 & 5.54 & 26.52 & 85.30 & 26.12& 4.03 & 24.37 & 83.16& 25.01 & 3.23& 22.58 & 82.56\\
    BertABS (\citeyear{liu2019text}) & 31.56& 5.02& 28.05& -& 27.42& 4.88& 25.15 & -& 25.45 & 3.82 & 23.04& -\\
    PRIMERA (\citeyear{xiao2022primera})& 31.90 & 7.40 & - & - & 32.04&5.78 &28.32  & 83.93& 29.99& 5.07 & 26.10 & 84.26\\
    TAG (\citeyear{chen2022target})& - & - & - & -& 30.48 & 6.16 & 27.79 & - & 28.04 & 4.75 & 25.33 & -\\
    Pointer-Generator (\citeyear{see2017get}) & 34.11& 6.76& 30.63 & 85.15 & 31.70 & 6.41& 29.01 & 82.52 & 28.53 & 4.96& 25.78&  82.71\\
    $\rm{UR^3WG}$ (\citeyear{shi2023towards}) & - & - & - & -& 32.68 & \underline{7.74} & 28.87 & - & 31.59 & 5.86 & 26.13 & -\\
    EDITSum (\citeyear{WANG2023103409}) & \underline{37.23}& \underline{8}& \underline{32.82} & \underline{86.11}& \underline{35.44} & 7.54& \underline{31.99}& \underline{85.11} & \underline{32.36} & \underline{6.2}& \underline{28.89} & \underline{85.40}\\
    \midrule
   \multicolumn{13}{l}{\emph{GPT-3.5-turbo-instruct}}\\\midrule
 Zero-Shot&31.19&5.89&27.57 & 83.86 &32.64&5.59&29.22& 83.01&29.24&4.51&25.97& 82.71\\
 One-Shot+BM25&32.55&6.18&28.43& 84.82&31.59&5.35&28.23& 83.56&29.3&4.49&25.87 & 83.59\\
 One-Shot+SciBERT&32.74&6.24&28.56& 84.87&31.59&5.37&28.23& 83.56&29.37&4.48&25.95& 83.60 \\\midrule
 \midrule
 \rowcolor{gray!30} \multicolumn{13}{l}{\emph{Our Experiments --- Incorporating DIR into Different Base Models ($k=3$, GPT-3.5-turbo-instruct as summary candidates) }}\\
 \midrule
  \rowcolor{gray!30} TransS2S (\citeyear{wang2022multi}) &35.92& 7.32& 31.46& 85.88 & 33.19& 6.74& 30.3& 83.03& 30.35& 5.36& 27.17& 83.42 \\
  \rowcolor{gray!30} TransS2S + DIR (Oracle)&37.56& 8.54& 32.9& 86.14 & 36.12& 7.78& 32.78& 85.03 & 33.05& 6.33& 29.37& 85.10 \\
 \rowcolor{gray!30} TransS2S + DIR (Pred)&\textbf{37.27}\tnote{$\dagger$} & \textbf{8.4}\tnote{$\dagger$}& \textbf{32.64}\tnote{$\dagger$}& \textbf{86.10}\tnote{$\dagger\dagger$} & \textbf{35.69}\tnote{$\dagger$}& \textbf{7.59}\tnote{$\dagger$}& \textbf{32.35}\tnote{$\dagger$}&  \textbf{84.98}\tnote{$\dagger$} & \textbf{32.61}\tnote{$\dagger$}& \textbf{6.17}\tnote{$\dagger$}& \textbf{29}\tnote{$\dagger$}& \textbf{85.03}\tnote{$\dagger$}\\ \midrule
 \rowcolor{gray!10} KGSum (\citeyear{wang2022multi})&36.65& 7.84& 31.9& 86.04 & 34.89& 7.22& 31.11& 84.63& 31.41& 5.72& 27.72& 84.84\\
\rowcolor{gray!10}  KGSum + DIR (Oracle)&38.06& 8.98& 33.04& 86.37& 36.54& 8.11& 33.1& 85.30& 33.24& 6.27& 29.91& 85.41\\
 \rowcolor{gray!10} KGSum + DIR (Pred)&\textbf{37.55}\tnote{$\dagger$}& \textbf{8.73}\tnote{$\dagger$}& \textbf{32.73}\tnote{$\dagger$}& \textbf{86.16}& \textbf{36.05}\tnote{$\dagger$}& \textbf{7.87}\tnote{$\dagger$}& \textbf{32.71}\tnote{$\dagger$}& \textbf{85.25}\tnote{$\dagger$}& \textbf{32.74}\tnote{$\dagger$}& \textbf{6.06}\tnote{$\dagger\dagger$}& \textbf{29.41}\tnote{$\dagger$}& \textbf{85.34}\tnote{$\dagger\dagger$}\\ \midrule
 \rowcolor{gray!30} BART (\citeyear{lewis2020bart})&32.83& 6.36& 28.36& 85.12& 31.16& 5.63& 27.91& 83.28& 29.69& 5.02& 26.18& 84.05\\
  \rowcolor{gray!30} BART + DIR (Oracle)&33.48& 6.88& 29.37& 85.28 & 32.77& 6.15& 29.52& 84.05& 30.29& 5.2& 26.72& 84.23\\
 \rowcolor{gray!30} BART + DIR (Pred)&\textbf{33.42}\tnote{$\dagger\dagger$}& \textbf{6.82}\tnote{$\dagger\dagger$}& \textbf{29.31}\tnote{$\dagger\dagger$}& \textbf{85.22}& \textbf{32.73}\tnote{$\dagger$}& \textbf{6.22}\tnote{$\dagger\dagger$}& \textbf{29.49}\tnote{$\dagger$}& \textbf{84.03}\tnote{$\dagger$}& \textbf{30.21}\tnote{$\dagger\dagger$}& \textbf{5.14}& \textbf{26.68}\tnote{$\dagger\dagger$}& \textbf{84.20}\\
    \bottomrule
    \end{tabular}
    \vspace*{-0.5\baselineskip}
    \end{threeparttable}
\end{table*}

\textit{3) Large Language Model (LLM)}: We use GPT-3.5-turbo-instruct\footnote{\url{https://platform.openai.com/docs/models/gpt-3-5}} as the representative LLM to explore the capability of LLM for MDSS with prompt engineering. We consider three different strategies to prompt GPT-3.5-turbo-instruct: \underline{Zero-Shot}: We use natural language instructions to prompt LLM without demonstrations. \underline{One-Shot+BM25}: We
prompt the LLM with instructions and one demonstration, which is retrieved by the classical sparse retrieval method BM25 \citep{robertson2009probabilistic} from the training set.  \underline{One-Shot+SciBERT}: We use SciBERT \citep{beltagy2019scibert}  to encode the examples and retrieve the most similar demonstration from the training set.

\vspace{-0.5em}
\subsection{Experimental Details}

\subsubsection{Parameter Settings}
We conduct experiments on three models: TransS2S \citep{wang2022multi}, KGSum \citep{wang2022multi}, and BART \citep{lewis2020bart}. All model-specific hyper-parameters are the same as in the original papers. We only set the hyper-parameters related to DIR. Specifically, both the loss coefficients $\lambda$ and $\beta$ are set to 0.01. We employ KL annealing \citep{bowman2016generating} to alleviate the degeneration of CVAE, gradually increasing
the KL multiplier from 0 to 1 over 100,000 steps. The batch size is set to 4 for TransS2S and KGSum, and set to 1 for BART. And the total training steps are 80,000 for TransS2S and KGSum, and 180,000 for BART. During decoding, we set the minimal decoding length to 110 and the maximum decoding length to 200. We also apply beam search with beam size 5 and length penalty \citep{wu2016google} with factor 0.4.

\subsubsection{First-Stage Summary Candidates Generation}
\label{sec53}
We consider three types of first-stage models to produce the summary candidates: Large Language Model, KGSum \citep{wang2022multi}, and BART \citep{lewis2020bart}. We use each type of model to generate \textbf{six} candidates.

For LLM, we use GPT-3.5-turbo-instruct to generate the candidates with six different prompt strategies, including 1 zero-shot prompting and 5 one-shot prompting with different demonstration retrieval methods. For KGSum and BART, we fully train them and use six different training checkpoints to obtain six summary candidates for each instance. 

\vspace*{-1\baselineskip}
\subsection{Experimental Results}
\subsubsection{Effectiveness of DIR --- Automatic Evaluation}
We incorporate our DIR framework into three different types of representative Transformer-based MDSS models: TransS2S \citep{wang2022multi}, KGSum \citep{wang2022multi}, and BART \citep{lewis2020bart}. We demonstrate the results in Table \ref{autoresult}. It lists the results of several baselines and three groups of our experiments, each of which includes the performance of the base model and DIR-equipped model (with oracle ranking and predicted ranking). In this experiment, we use the outputs from GPT-3.5-turbo-instruct as summary candidates.


We observe that our DIR can significantly improve the performance of base models. For instance, on TransS2S, DIR leads to an increase of 3.76\%, 14.75\%, and 3.75\% for R-1, R-2, and R-L on Multi-Xscience dataset, an average of 7.42\% improvement. On TAD and TAS2 datasets, the average improvements are 8.97\% and 9.76\%. A similar trend of improvements can also be found on KGSum and BART, where the average improvements on ROUGE metrics across the three datasets are 5.57\% and 4.40\%. The results on BERTScore also consistently demonstrate the effectiveness of our DIR. Furthermore, when looking at \textbf{KGSum + DIR (Pred)}, we observe that our model achieves a new state-of-the-art on the three datasets, compared with the previous SOTA model EDITSum \citep{wang2023plan}. It is worth noting that, KGSum + DIR (Pred) is not the best-performing model in all experiments, as we will show in subsequent results.

\vspace{-0.3em}
\subsubsection{Effectiveness of DIR --- Human Evaluation}
\label{sec5.4.2}
We also conduct a human evaluation to assess the effectiveness of DIR. We set up nine sets of comparison experiments on: three Transformer-based models (TransS2S, KGSum, BART) $\times$ three datasets (Multi-Xscience, TAD, TAS2). In each set, we randomly select 50 summaries for the base model and the DIR-equipped model, respectively. We invite three graduate students with expertise in natural language processing to assess the summaries. The assessors are shown the source papers and the two summaries, and they need to judge which one is better, or choose a tie if the two are of equal quality, according to three essential aspects: \emph{Informativeness}, \emph{Fluency}, and \emph{Succinctness}.

\begin{figure}[h]
  \centering
  \vspace*{-0.4\baselineskip}
   \includegraphics[width=0.95\columnwidth]{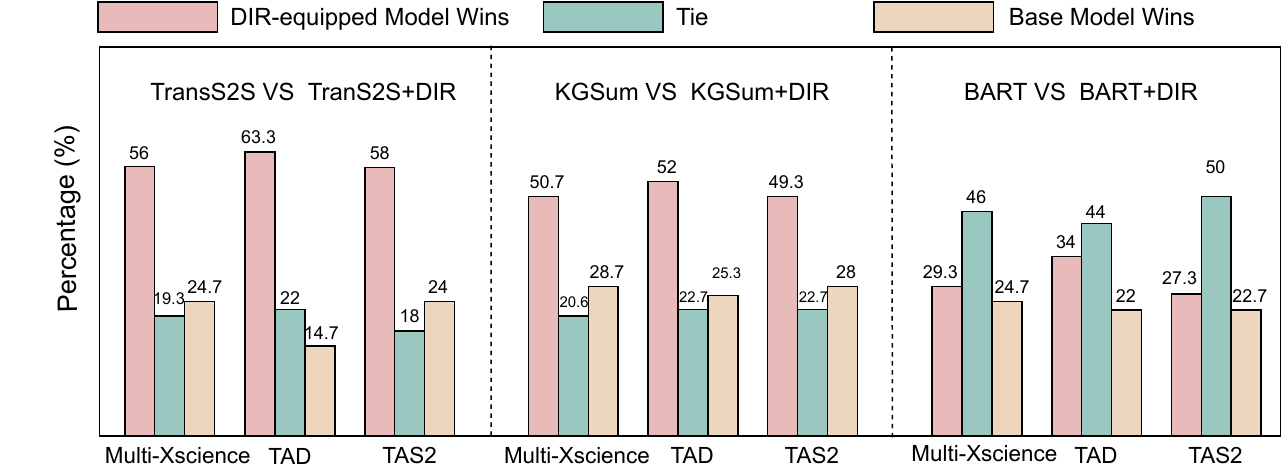}
    \vspace*{-0.2\baselineskip}
     \caption{Human evaluation results for three groups of models across three datasets.}
     \vspace{-0.8em}
    \label{humaneval}
\end{figure}

The result is shown in Figure \ref{humaneval}. We observe a clear preference for DIR-equipped models compared with the base models. Specifically, incorporating DIR into TransS2S obtains an average of 59.1\% preference on the three datasets, whereas the average preference of the base model is 21.13\%. Similarly, the average preference of KGSum+DIR is 50.67\% on the three datasets, compared to 27.33\% of the base model. On BART model, the assessors choose a tie for nearly half of the summaries, while the average preference of DIR-equipped model is 30.2\%, slightly higher than the base model's 23.13\%. The human evaluation results indicate the effectiveness of our DIR framework in improving the performance of different summarization models.

\vspace{-0.5em}
\subsubsection{Impact of Candidate Type and Number} We also conduct experiments to explore to what extent is the performance of DIR-equipped model influenced by candidate type and number? 

\noindent\textbf{Impact of Candidate Type.} We first experiment on candidate type. For this purpose, we consider three types of summary candidates (GPT-3.5-turbo-instruct, BART, and KGSum, described in \S{\ref{sec53}}). Due to the limitation of training cost, we only experiment on Multi-Xscience dataset, where we incorporate the three types of candidates into two Transformer-based MDSS models (TransS2S and KGSum). The result is shown in Figure \ref{sumcand_eval}.

\begin{figure}[th]
  \centering
   \includegraphics[width=0.95\columnwidth]{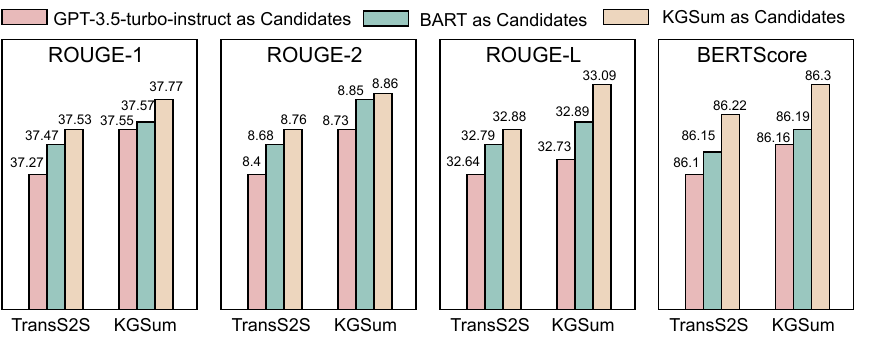}
   \vspace{-0.5em}
     \caption{The performance of DIR-equipped TransS2S and KGSum when using different types of summary candidates.}
     \vspace{-0.6em}
    \label{sumcand_eval}
    \vspace*{-0.6\baselineskip}
\end{figure}

We observe that different types of candidates consistently lead to performance gains in DIR-equipped models. Generally, higher-quality summary candidates bring better performance (As listed in Table \ref{autoresult}, KGSum > BART > GPT-3.5-turbo-instruct according to candidate quality.). Among all the results in Figure \ref{sumcand_eval}, incorporating DIR into KGSum with the outputs from KGSum as candidates achieves the top performance, reaching the brand-new SOTA. However, we should point out that, although GPT-3.5-turbo-instruct is inferior to BART and KGSum, it does not require additional training to generate candidates. In contrast, BART and KGSum need to be fully trained on MDSS datasets to generate candidates. \textit{Therefore, the summary candidates-based summarization paradigm becomes more promising in this era of large language models, because high-quality candidates can be obtained at little cost.}

\noindent\textbf{Impact of Candidate Number }$\bm{k}$. We then experiment on candidate number. For this purpose, we consider the number of selected summary candidates $k \in \{ 1,2,3,4,5,6\}$. The experiment is based on TransS2S model with the outputs from GPT-3.5-turbo-instruct as summary candidates, and is conducted on Multi-Xscience dataset. The result is shown in Figure \ref{number}. 

\begin{figure}[th]
  \centering
  \vspace*{-0.5\baselineskip}
   \includegraphics[width=0.95\columnwidth]{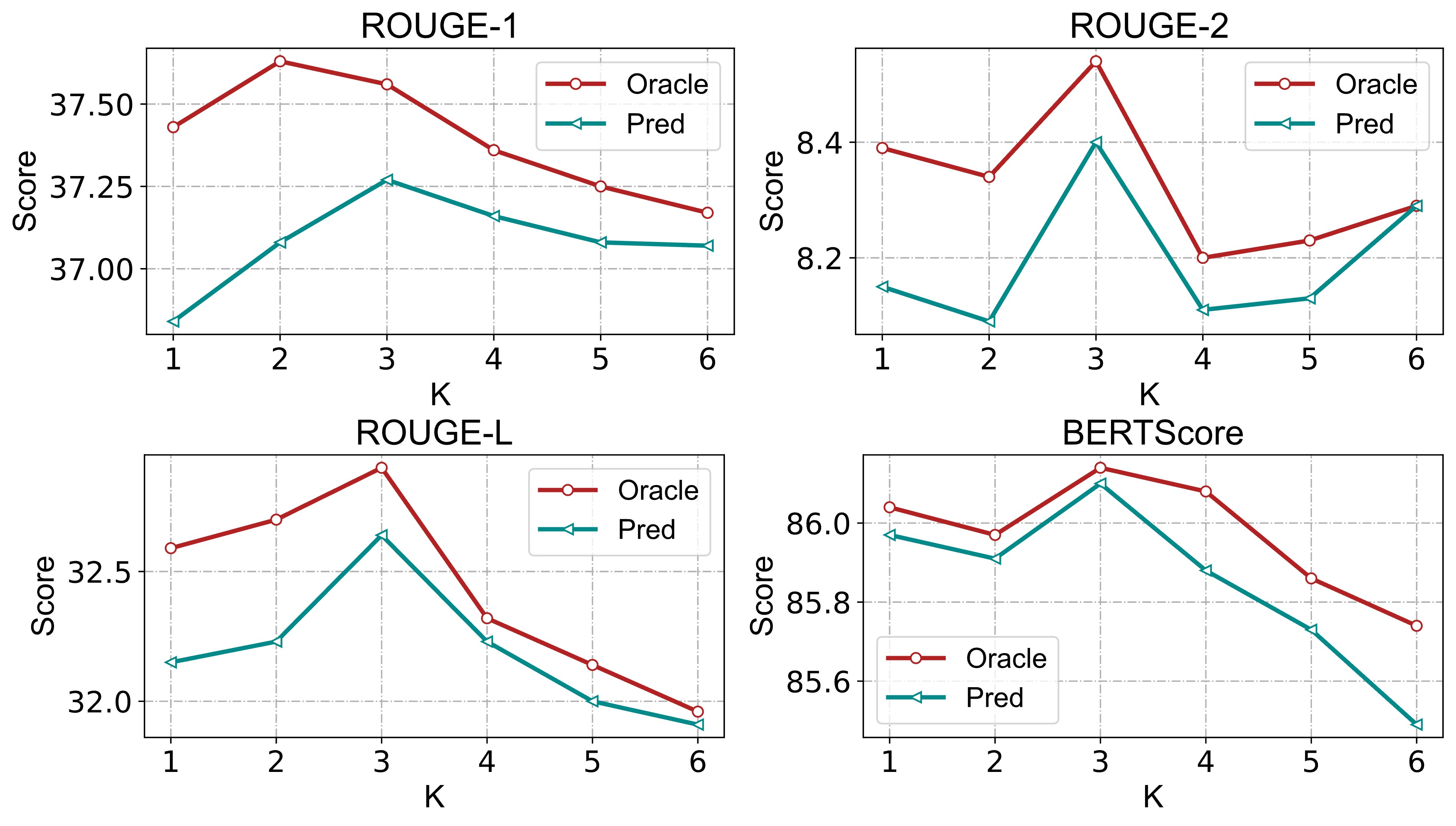}
   \vspace{-1em}
     \caption{The performance of different numbers of selected summary candidates $k$. }
     \vspace{-0.7em}
    \label{number}
\end{figure}

We can obtain three observations from Figure \ref{number}. Firstly, the number of selected candidates $k$ has a great impact on model performance. When $k=3$, the performance on ROUGE and BERTScore reaches the peak. Secondly, incorporating more candidates into summarization cannot always lead to better performance. Actually, we can see from Figure \ref{number} that using all the 6 candidates gets the worst performance. The reason is, while we have found from \S{\ref{sec3.2}} that more candidates bring more positive information, the concatenation of multiple candidates makes the candidates longer than source papers, thus weakening the role of source papers. \textit{This observation also validates our motivation that selecting higher-quality candidates, rather than using the entire candidates set, helps to produce better summaries. }Thirdly, the gap in ROUGE scores between \textit{Oracle} and \textit{Pred} is the largest when $k=1$, and the gap decreases when $k$ increases. This is because the \textit{Pred} performance of DIR-equipped models is directly influenced by the predicted ranking results from the Candidates Ranking Module (\S{\ref{sec_cr}}). When $k=1$, the predicted top-1 ranking can be completely different from the oracle top-1 ranking, resulting in the huge performance gap.

\vspace{-0.5em}
\subsection{More Analyses on DIR}
\subsubsection{Ablation Study}
Our DIR framework consists of several key components, including candidates ranking (\S{\ref{sec_cr}}), information modeling with latent variables (\S{\ref{sec_iird}}), and information disentangling ((\S{\ref{sec_iird}})). To investigate how different components affect the performance, we conduct an ablation study on TransS2S model with the outputs from GPT-3.5-turbo-instruct as summary candidates and $k=3$. Table \ref{ablation} shows the results on Multi-Xscience dataset. Next, we give a detailed discussion about each component:

\begin{table}[th]
\vspace*{-0.9\baselineskip}
    \caption{Ablation study results on Multi-Xscience dataset.}
    \label{ablation}
    \centering
    \small
    \vspace{-0.7em}
    \renewcommand\arraystretch{0.8}
    \begin{tabular}{m{0.16\textwidth}<{\raggedright}|p{0.85cm}<{\centering}p{0.85cm}<{\centering}p{0.87cm}<{\centering}p{0.8cm}<{\centering}}
    \toprule
   \textbf{Model} & \textbf{R-1 (\%)} & \textbf{R-2 (\%)} & \textbf{R-L (\%)} & \textbf{BS (\%)} \\ 
   \midrule
   \textbf{TransS2S+DIR} & \textbf{37.27} & \textbf{8.4} & \textbf{32.64} & \textbf{86.10}\\   
   w/o-CandRank & 36.91 & 8.03 & 32.17 & 85.83 \\
   w/o-LatVar & 36.52 & 7.9 & 31.95& 85.91 \\
   w/o-InfoDis & 37.07 & 8.27 & 32.39 & 86.01 \\
    \bottomrule
    \end{tabular}
    \vspace*{-1\baselineskip}
\end{table}

\noindent$\bullet$ \textit{w/o-Candidates Ranking (w/o-CandRank)}: To assess the effectiveness of candidates ranking, we replace the ranking module with random selection. Specifically, we train a new model with candidates random selection and test it in the same setting. From Table \ref{ablation}, we observe the performance boost from the candidates ranking module, which verifies our motivation that selecting higher-quality candidates contributes to generating better summaries.

\noindent$\bullet$ \textit{w/o-Latent Variables (w/o-LatVar)}: To assess the effectiveness of our instructive information modeling with latent variables, we remove the whole instructive information modeling and disentangling module (\S{\ref{sec_iird}}), leaving only the concatenation of source context and candidate context for decoding. From Table \ref{ablation}, we observe that removing the latent variables modeling of candidate information leads to performance degradation. This demonstrates the effectiveness of our information modeling with latent variables.

\noindent$\bullet$ \textit{w/o-Information Disentangling (w/o-InfoDis)}: To assess whether it is effective to disentangle information into positive and negative aspects, we model candidate information into a latent variable without distinguishing positive and negative information. The result reveals that our information disentangling module can help improve summarization performance.

\vspace*{-0.5\baselineskip}
\subsubsection{Analysis of Disentanglement} An important module of our DIR framework is the separation of positive and negative information. To illustrate whether the information is disentangled, we visualize the positive and negative information before and after disentangling training with t-SNE in Figure \ref{tsne}. We obtain latent variables for the whole test set of Multi-Xscience for visualization. From Figure \ref{tsne}, we can find that positive and negative information can be well separated after disentangling training, which suggests that the employed informativeness objective and adversarial objective are conducive to learning disentangled representations.

\begin{figure}[th]
  \centering
  \vspace*{-0.3\baselineskip}
   \includegraphics[width=0.9\columnwidth]{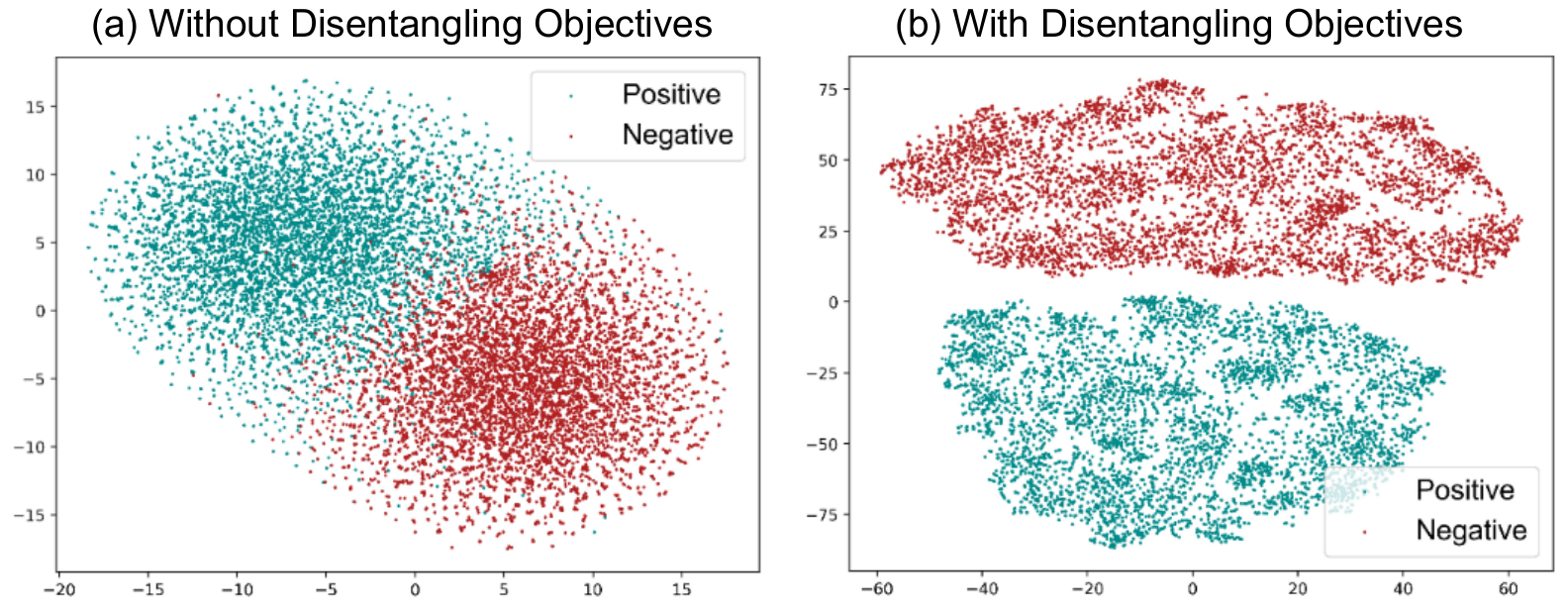}
   \vspace*{-0.7\baselineskip}
     \caption{Visualization of positive and negative latent variables before and after disentangling training.}
     \vspace*{-0.7\baselineskip}
    \label{tsne}
    \vspace*{-0.5\baselineskip}
\end{figure}

\vspace{-0.5em}
\subsubsection{Case Study}
We also provide a case study to illustrate how our 
DIR affects summary generation in Figure \ref{case}. In Figure \ref{case}, positive information is marked with \colorbox{Salmon!60}{pink shading}, while negative information is marked with \colorbox{PineGreen!40}{green shading}. As observed from Figure \ref{case}, DIR can effectively enhance the influence of positive information during decoding by disentangling positive and negative information and integrating them into the decoder. This results in the appearance of positive information such as ``\textit{hypothesis testing}'', ``\textit{community evolution}'', and ``\textit{state dynamics}'' in the summary, while avoiding negative information like ``\textit{time-evolving systems}'', ``\textit{high-resolution time-varying}'', and ``\textit{discrete states}''. In contrast, TransS2S, which only takes input documents for decoding, lacks the ability to utilize external guidance information for summary generation, which is prone to generate contents that mismatch the gold summary. This case illustrates that the proposed DIR can effectively mitigates the issue of uncontrollable content during decoding by modeling guidance information from summary candidates as positive and negative information and integrating them into the decoder.

\begin{figure}[th]
  \centering
  \vspace{-0.5em}
   \includegraphics[width=0.9\columnwidth]{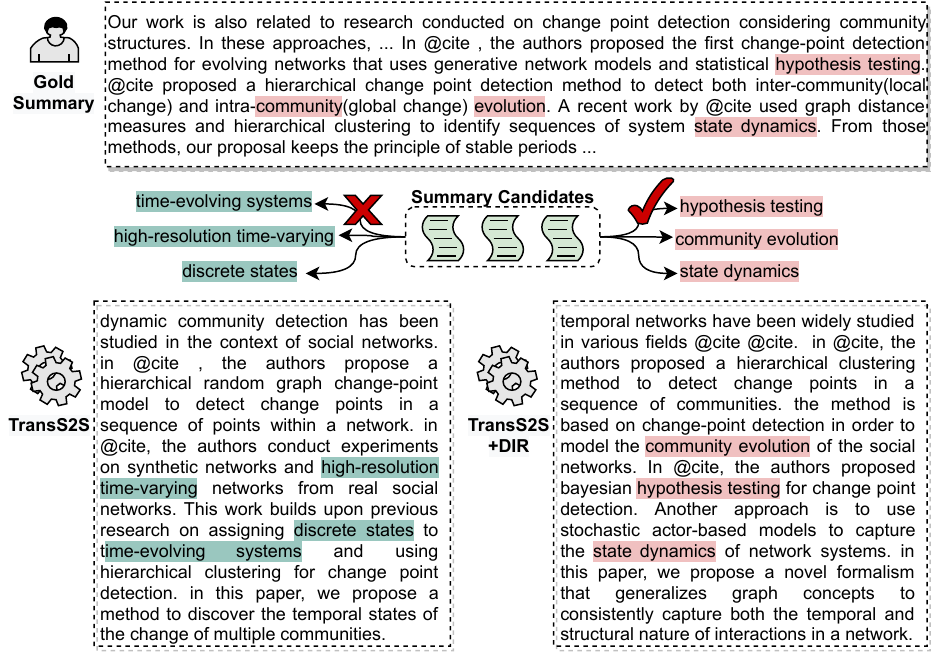}
   \vspace*{-0.8\baselineskip}
     \caption{Case Study: Comparison between summaries generated by TransS2S and DIR-equipped TransS2S.}
    \label{case}
    \vspace{-0.8em}
\end{figure}

\vspace{-1em}
\section{CONCLUSION AND FUTURE WORK}
In this paper, we propose a summary candidates fusion framework --- DIR, to enhance the decoder's handling of global information and control over summary content for MDSS. DIR is compatible with various Transformer-based MDSS models, which consists of three components: candidates ranking module, information disentangling module, and information fusion module. We conduct extensive experiments on three types of MDSS models with three types of summary candidates across three MDSS datasets. Experimental results verify the superiority of our DIR framework in improving the performance of summarization. More analyses also demonstrate the rationality of the proposed individual modules. In this era of large language models, we believe our method will become more promising with the assistance of low-cost, high-quality candidates. 

\section*{Acknowledgments}
This work was supported by the National Key Research and Development Project of China (No. 2021ZD0110700) and Hunan Provincial Natural Science Foundation (Grant Nos. 2022JJ30668). The authors would like to thank the anonynous reviewers for their valuable comments and suggestions to improve this paper.
\bibliographystyle{ACM-Reference-Format}
\bibliography{my_reference}










\end{document}